\documentclass[conference]{IEEEtran}
\IEEEoverridecommandlockouts

\usepackage{cite}
\usepackage{amsmath,amssymb,amsfonts}
\usepackage{algorithmic}
\usepackage{graphicx}
\usepackage{textcomp}
\usepackage{xcolor}
\usepackage{hyperref}

\def\BibTeX{{\rm B\kern-.05em{\sc i\kern-.025em b}\kern-.08em
    T\kern-.1667em\lower.7ex\hbox{E}\kern-.125emX}}
\begin{document}

\title{Improved intent classification based on context information using a windows-based approach 
\\
}

\author{
\IEEEauthorblockN{Jeanfranco D. Farfan-Escobedo}
\IEEEauthorblockA{Institute of Computing\\University of Campinas\\
Campinas/SP, Brazil\\
Email: j230222@dac.unicamp.br}
\and
\IEEEauthorblockN{Julio C. Dos Reis}
\IEEEauthorblockA{Institute of Computing\\University of Campinas\\
Campinas/SP, Brazil\\
Email: dosreis@unicamp.br}
}

\maketitle

\begin{abstract}

Conversational systems have a Natural Language Understanding (NLU) module. In this module, there is a task known as an intent classification that aims at identifying what a user is attempting to achieve from an utterance. Previous works use only the current utterance to predict the intent of a given query and they do not consider the role of the context (one or a few previous utterances) in the dialog flow for this task. In this work, we propose several approaches to investigate the role of contextual information for the intent classification task. Each approach is used to carry out a concatenation between the dialogue history and the current utterance. Our intent classification method is based on a convolutional neural network that obtains effective vector representations from BERT to perform accurate intent classification using an approach window-based. Our experiments were carried out on a real-world Brazilian Portuguese corpus with dialog flows provided by Wavy global company. Our results achieved substantial improvements over the baseline, isolated utterances (without context), in three approaches using the user's utterance and system's response from previous messages as dialogue context.
 
\end{abstract}


\section{Introduction}
\label{sec:Introduction}
When a human-to-human conversation takes place, people “deduce” intention of a sentence based on the context of the conversation (one or a few previous utterances). In this sense, people do not interpret an intention-based only on an isolated utterance within a dialogue flow. Consequently, the analysis of an utterance to identify a user’s intention in a dialogue system can benefit from the conversation context. However, literature on intent classification does not address this significant approach.

Within a dialogue system, two agents widely interact during the communicative process: \textit{user} and \textit{system}. A conversation is usually structured in turns; each turn is defined by one utterance from the user and single system response. As a result of this process of interaction between system and user, the set of turns forms a dialog flow.  


Table \ref{tab:Conversational_flow} shows an example of a conversation between a user and system, in which $U$ represents the user and $S$ system, The conversational flow column represents the intent for each utterance $U$.

\begin{table}[h!]
\scriptsize	
\caption{An example of a conversation between an user and system.}
\begin{center}
\begin{tabular}{p{6.5cm}||p{2cm} } 
 \hline
 Utterance & Conversational flow \\ 
 \hline
 U: I’d like Peruvian food & food\_information\\ 
 S: "La Clave del Sabor" is a delicious restaurant located near the historic downtown of Cusco. &  \\ 
 \hline
 \textcolor{red}{U: But, I am allergic to shrimps} & user\_information\\
 S: Don't worry, "La Clave del Sabor" has beef, chicken, alpaca, guinea pig and fish food & \\
 \hline
 \textcolor{red}{U: Is it reasonably priced?} & cost\_information\\ 
 S: Yes, "La Clave del Sabor" is in the moderate price range & \\
 \hline
 U: What is the phone number? & request\_information \\ 
 S: The number of "La Clave del Sabor" is +51974264215. & \\ 
 \hline
\textcolor{red}{U: I need to travel the next day to Machu Picchu, is there a tour agency nearby?} & another\_information \\
 S: Two minutes from "La Clave del Sabor" restaurant is the "Inca Travel" tourism agency & \\ 
 \hline
\end{tabular}
\end{center}
\label{tab:Conversational_flow}
\end{table}

Literature has not carried out comprehensive studies using the context in the intent classification task. To the best of our knowledge, there is a lack in how to solve this problem. Currently, studies in literature do not examine the relevance of context during intent classification task only use the current utterance as input to predict the user's intent \cite{chen2019bert, zhang2019ensemble, shridhar2018subword}.  As for other tasks such as Dialogue State Tracking and Dialogue Acts do not standardize the context used. Some works used all previous information, and others used the last system reply.

We propose a novel approach to managing context, a windows-based procedure. Our strategy aims to make use of windows on the previous information of the dialog flow.

In our work, we proposed several combinations to use a dialog flow’s earlier conversations to find the best contextual information. These approaches use previous utterances of a dialog flow to increase the contextual knowledge to improve our model. This architecture makes it possible to train the dialogue history and the current utterance jointly. Our approach is robust to imbalanced datasets due to we modify our loss function to penalize misclassification of the underrepresented classes more than the dominant ones. Our results achieved substantial improvements over the baseline in three approaches using the user's utterance and system's response from previous messages as dialogue context.

The remainder of this paper is organized as follows. Section~\ref{sec:RELATED_WORKS} presents related works. Section~\ref{sec:METHODOLOGY} makes known our methodology. Section~\ref{sec:RESULTS} presents the experimental results and discusses our findings. Section~\ref{sec:CONCLUSIONS_FUTURE_WORKS} concludes and suggests future works.

\section{RELATED WORKS}
\label{sec:RELATED_WORKS}
We examined different existing investigations, and they were organized into three categories. First, we reviewed recent papers, who address the intent classification task-based in isolated utterance and disconnected sentences. Chen \textit{et al.}~\cite{chen2019bert} proposed a joint intent classification and slot-filling model based on BERT, aiming at addressing the poor generalization capability of traditional NLU models. Shridhar \textit{et al.} ~\cite{shridhar2018subword} used Semantic Hashing as embedding for the task of Intent Classification on three frequently used benchmarks: AskUbuntu, Chatbot and Web Application. Similarly, Active Learning methods were provided to deal with this task. Zhang \textit{et al.}~\cite{zhang2019ensemble} designed an ensemble deep active learning method, which constructs intent classifiers based on BERT and uses an ensemble sampling method to choose informative data for efficient training in Chinese and English languages. Farfan \textit{et al.}~\cite{farfan2021active} analyzed active learning techniques minimized the amount of labeled data required to build prediction models.

Second, we review whether the current tools use the context to the intent classification task. According to Liu \textit{et al.}~\cite{liu2021benchmarking}, none of the publicly available Natural Language Understanding (NLU) toolkits, such as Dialogflow, LUIS and Rasa, use dialogue context for Intent classification and NER.

Third, we analyzed proposals that use dialog flow in other domains. Several types of research have explored dialog flow to improve the accuracy in different tasks. Khanpour \textit{et al.}~\cite{khanpour2016dialogue} applied a deep-stacked Long Short-Term Memory (LSTM) with pre-trained word embeddings to classify dialogue acts (DAs) in open-domain conversations. The main reason for stacking LSTM cells is to gain longer dependencies between terms in the input chain. As far as dialogue state tracking task is concerned, Gulyaev \textit{et al.}~\cite{gulyaev2020goal} proposed a GOaL-Oriented Multi-task BERT-based dialogue state tracker (GOLOMB) inspired by architectures for reading comprehension question answering systems. The model uses dialogue history to predict the next slots. Similarly, Wu \textit{et al.}~\cite{wu2020tod} proposed task-oriented dialogue BERT (TOD-BERT). This pre-trained model outperforms strong baselines like BERT on several downstream task-oriented dialogue applications. TOD-BERT concatenates all the utterances in the same dialogue into one to capture speaker information and the underlying interaction behavior in dialogue. In the same way, Chao~\cite{chao2019bert} proposed BERT-DST, an end-to-end dialogue state tracker which directly extracts slot values from the dialogue context. BERT-DST used BERT to identify slot values from their semantic context. Also, they used the system utterance from the previous turn and the current turn user utterance as dialogue context input.

Literature has presented solutions based only on isolated utterances for this tasks. Also, solutions without a standard to select the context for dialogue acts, and dialogue state tracking tasks. Therefore, there is a lack of studies addressing the context relevance and the most relevant previous information for intent classification task and other domains.

\section{METHODOLOGY}
\label{sec:METHODOLOGY}
\begin{figure*}[h]
  \includegraphics[width=\textwidth,height=10cm]{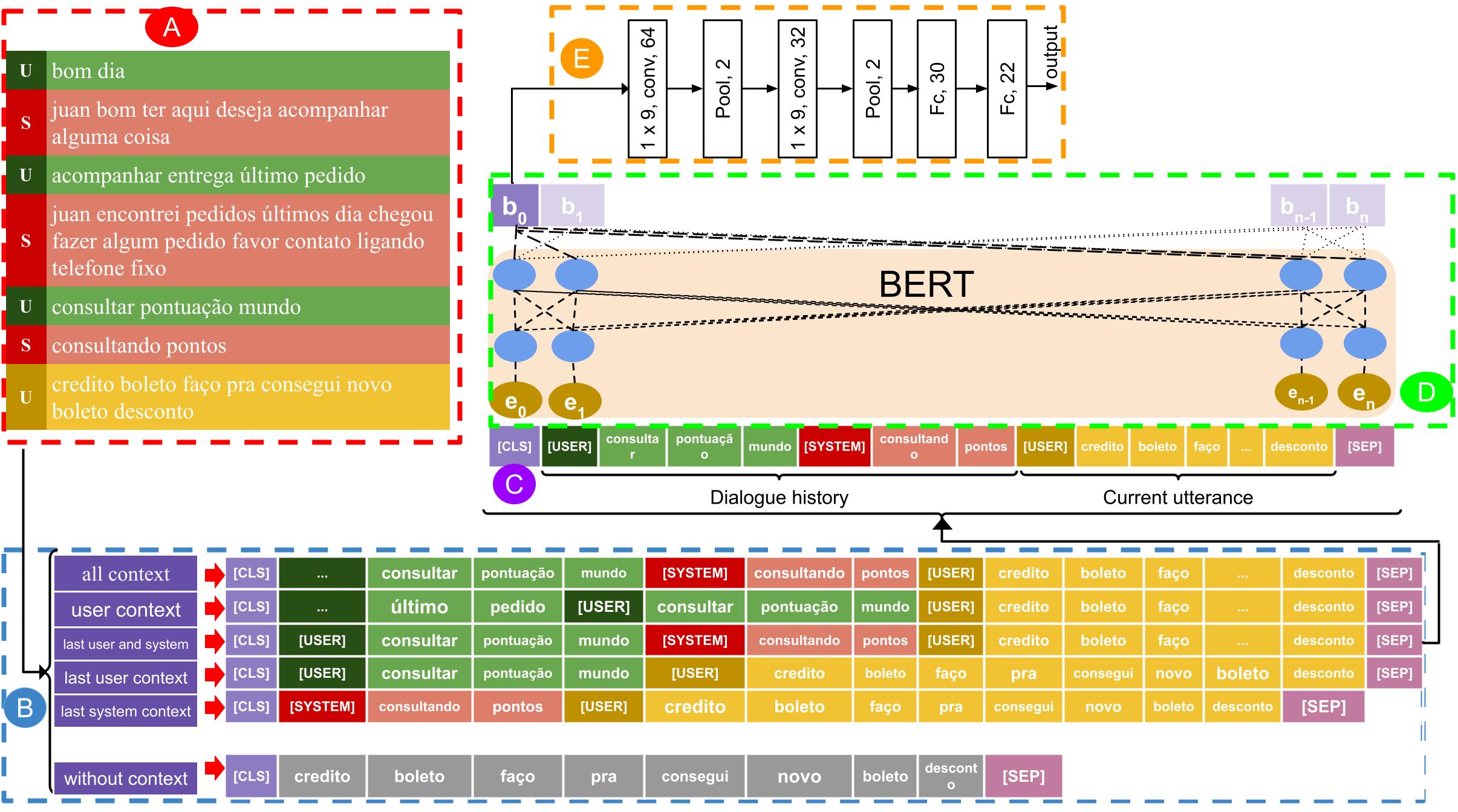}
  \caption{We defined five dialogue representation trajectories: 'all context', 'user context', 'last user and system', 'last user context', and 'last system context'. Each representation is used to carry out a concatenation between the dialogue history and the current utterance. 'all context’ uses the complete previous information (user and system) like context; ’user context’ employs the user’s previous messages to get the context; 'last user and system' return the context using the user and system's last utterances; 'last-user context' employs the user’s last utterance to get the context. 'last-system context’ makes use of the system’s last utterance to get the context.}
  \label{fig:methodology}
\end{figure*}

We proposed a novel approach to managing context, a windows-based procedure. Our window-based strategy aims to make use of windows on the previous information of the dialog flow. Figure \ref{fig:methodology} presents our methodology. It is mainly composed of four components (dotted lines of different colors delimited each module): preprocessing (dotted red lines); dialogue context module (dotted blue lines); feature extraction (dotted green lines); and classifier (dotted orange lines).

Figure \ref{fig:methodology} shows a dialog flow pre-processed (red circle, A), then presents several types of trajectories encoded in our solution (blue circle, B). Our model takes a dialogue context and the current utterance as input for each user turn (purple circle, C). The BERT-based encoding module encodes the dialogue context input to produce contextualized sentence-level and token-level representations (green circle, D). The classification module then uses the sentence-level representation ($b_{0}$) to generate a categorical distribution over twenty-two types of categories (orange circle, E). 
 
\subsection{Preprocessing}
We applied to preprocess to clean input queries. The preprocessing step includes lower-casing, removing punctuation on the corpus, removing URLs, removing stop words and the message is finally a vector of tokens, for instance: For the utterance ’Já paguei o boleto da campanha \#\# porque ele ainda está pendente no site https://campanha.com/5cb4abba34070929d959d32d?’ its correspondent output is ’paguei boleto campanha porque ainda pendente’.

\subsection{Dialogue context module}
To capture user information and the system's response in a dialogue flow, we used two special tokens, $[USER]$ and $[SYSTEM]$. We prefix the special tokens to each user utterance and system response, then concatenate all utterances in the same dialogue into one flat sequence (Figure \ref{fig:methodology}). For example, for a dialogue $D$ $=$ $\{U_{1}, S_{1}, . . . , U_{n}, S_{n}\}$, where $n$ is the number of conversation turns (messages send by the user), and each $U_{i}$ or $S_{i}$ contains a sequence of words, the input of the pre-training model is processed as ``$[USER]$ $U_{1}$ $[SYSTEM]$ $S_{1}$ $...$'' with standard positional embeddings and segmentation embeddings. The final representation starts with a $[CLS]$ token followed by the dialogue representation, then the $[SEP]$ token is used for delimiting the representation so that it can be fed to the Feature extraction module.

\subsection{Feature extraction}
The feature extraction module is based on BERT \cite{devlin2018bert}. Figure \ref{fig:methodology} presents the user utterance and system response from the previous turn as dialogue context (purple circle, C). Our solution also uses the current user utterance message as input, represented as a token sequence in BERT input format. The first token is $[CLS]$, followed by the tokenized user utterance and system response, then the current tokenized user utterance, and  $[SEP]$. For instance, in Figure \ref{fig:methodology} we use the following token sequence. '\textit{$[CLS]$ $[USER]$ consultar  pontuação mundo $[SYSTEM]$ consultando pontos [USER] credito boleto faço $...$  desconto $[SEP]$}'.

Formally, let $[x_{0}, x_{1}, ... , x_{n}]$ denoting the input token sequence, then BERT input layer embeds each token $x_{i}$ into an embedding $e_{i}$, which is the sum of three embeddings:

\begin{equation}
    \scriptsize BERTinput(x_{i}) = Etok(x_i) + Eseg(x_i) + Epos(x_i), e_i \in \mathbb{R}^d, \forall \ 0 \leq i \leq n
\end{equation}

where $Etok$ is the token embedding, $Eseg$ is the segment embeddings, and $Epos$ is the position embedding for each token $x_{i}$. The embedded input sequence $[e_{0}, ... , e_{n}]$ is then passed to BERT bidirectional Transformer encoder, whose final hidden states are denoted by $[b_{0}, ... , b_{n}]$ (Figure \ref{fig:methodology}). 

\begin{equation}
    \footnotesize [b_0, ... , b_n] = BiTransformer([e_0, ... , e_n]), b_i \in \mathbb{R}^d, \forall \ 0 \leq i \leq n
\end{equation}

Our solution uses a contextualized sentence-level representation output ($b_{0}$) as a final state corresponding to the $[CLS]$ token. 

\subsection{Classifier}
The input for classification module is sentence-level representation $b_{0}$ from the feature extraction module. The classification module predicts the value of $b_0$ to be one of the twenty-two categories using a Convolutional Neural Network (CNN) and softmax.

We used a CNN to capture intrinsic syntactic and semantic patterns from input sentences \cite{schwarz2020emet}. CNN consists of convolution, pooling, and activation layers. At each convolution layer, a set of kernels convolved a vector, which acts as filters. The pooling layer is composed of a max-pooling function, which reduces the vector size and increases the receptive field size. The activation layer adds nonlinearity to the neural network. In this case, it is usually a Rectified Linear Units (ReLU) that replaces negative inputs with 0 and keeps the positive inputs unchanged (Table \ref{table:table_config_cnn_context}).

 \begin{table}[htp!]
 \scriptsize
 \begin{center}
 \caption{\label{font-table} Convolutional Neural Network configuration.}
 \begin{tabular}{ c|| c ||c ||c ||c ||c || c}
   \hline
   \textbf{Type} & \textbf{Description} & \textbf{\#filters} & \textbf{filter size} & \textbf{stride} & \textbf{\#units} & \textbf{rate}\\
   \hline
   BN & Batch normalization & - & - & - &-&-\\\hline
   conv & 1D Convolution & 64 & (1 $\times$ 9) & 1 & -&-\\\hline
   Pool & Max-pooling & - & (1 $\times$ 2) & 2 & -&-\\\hline
   bn & Batch normalization & - & - & -&-&-\\\hline
   do & Dropout & - & - & -&-&0.6\\\hline
 
   conv & 1D Convolution & 32 & (1 $\times$ 9) & 1 & -&-\\\hline
   Pool & Max-pooling & - & (1 $\times$ 2) & 2 & -&-\\\hline
   bn & Batch normalization & - & - & -&-&-\\\hline
   do & Dropout & - & - & -&-&0.6\\\hline
 
   FC & Fully-connected & - & - & - & 30&-\\\hline
   bn & Batch normalization & - & - & -&-&-\\\hline
   do & Dropout & - & - & -&-&0.25\\\hline
   FC & Fully-connected & - & - & - & 22&-\\
 
   \hline
 \end{tabular}
 \label{table:table_config_cnn_context}
 \end{center}
 \end{table}

\section{EXPERIMENTAL RESULTS}
\label{sec:RESULTS}

\subsection{Dataset}
For the evaluation of the proposed approach, we use Wavy Global Dataset (WvGD), which is a dataset of conversations between humans and chatbots implemented in Brazilian Portuguese. The dataset contains 7574 conversations with at least three and at most fifteen turns, where each turn is a result of interaction between a user's query and a system's response. As far as the full number of utterances is concerned, there are 36,056 queries, and the number of categories is 22. Table \ref{table:estadisticas} shows summary corpus's statistics. 
\begin{table}[htp!]
\footnotesize
\begin{center}
\caption{\label{font-table} Corpus statistics for the WvGD dataset.}
\begin{tabular}{c|| c}
  \hline
  Average number of words per user's query & 3.75 \\
  \hline
  Standard deviation per user's query & 2.92 \\
  \hline
  Variance per user's query & 8.57 \\
  \hline
  Average number of words per system's response & 33.76 \\
  \hline
  Standard deviation per system's response & 30.65 \\
  \hline
  Variance per system's response & 939.48 \\
  \hline

\end{tabular}
\label{table:estadisticas}
\end{center}
\end{table}

\subsection{Experimental Settings}
In our experimental procedure, we randomly divided the WvGD into 60\% of samples for training the classifier, 20\% for validation, and 20\% for testing it. We used the pre-trained multilingual BERT model. We used their feature extraction effectiveness to apply our window-based approach.

About the convolutional neural network architecture configuration, the loss function used was a cross-entropy loss for the corresponding prediction target. We updated all layers in the model using RMSprop optimization and early stopping on the validation set. During training, we used 25\% and 60\% dropout rates, also batch normalization. Finally, the model was trained for 50 epochs.


\subsection{Results and Discussion}
Results in Table \ref{table:results_test} reveal that our approaches based on the last utterance from user and system \textit{(user-system context, last-user context, and last-system context)} outperformed the baseline model in terms of accuracy, recall, precision, and f1-score. In contrast, baseline model presented better results for two approaches based on all conversation, \textit{all context} and \textit{user context}. These results make sense because the dialogue is not just a sequence of independent utterances, but rather a collective action performed by the user and the system.

\begin{table}[htp!]
\footnotesize
\begin{center}
\caption{\label{font-table} Comparison between isolated queries (without context) and window-based approaches. The highest results are highlighted in red. Baseline results are highlighted in blue.}

\begin{tabular}{ c ||c||c||c||c }
  \hline
    \textbf{Approach}&\textbf{Accuracy}&\textbf{Recall}&\textbf{Precision}&\textbf{F1-Score}\\
  \hline
  \textbf{without context} & \textcolor{blue}{85.32} & \textcolor{blue}{83.15} & \textcolor{blue}{87.55} & \textcolor{blue}{85.25} \\
  \hline
  \textbf{all context} & 66.52 & 55.22 & 80.54 & 65.14 \\
  \hline
  \textbf{user context} & 78.00 & 69.34 & 88.91 & 77.57 \\
  \hline
  \textbf{user-system context} & 86.08 & 82.94 & \textcolor{red}{\textbf{90.55}} & 86.49 \\
  \hline
  \textbf{last-user context} &  \textcolor{red}{\textbf{87.18}} & \textcolor{red}{\textbf{85.01}} & 89.79 & \textcolor{red}{\textbf{87.28}} \\
  \hline
  \textbf{last-system context} & 86.56 & 84.18 & 89.85 & 86.88 \\
  \hline
\end{tabular}
\label{table:results_test}
\end{center}
\end{table}

Obtained results do not consider dealing with imbalanced datasets, in which some of the classes appear much more often in the dataset than others. The problem is that the model is likely to learn to predict only the dominant classes. In this context, a potential
strategy is to modify our loss function to penalize misclassification of the underrepresented classes more than the dominant ones. We modify the cross-entropy loss function adding the loss value for each label. 
\begin{equation}\label{eq:1}
L_{CE} = - \sum_{i=1}^{n} t_i \log (P_i) \times L_i \textit{, for n classes,}
\end{equation}

where $t_{i}$ is the truth label, $P_{i}$ is the softmax probability and $L_{i}$ is the loss value for the $i^{th}$ class. As a consequence, adding loss values implies that we need to re-train our model, tunning its penalization during training. Table \ref{table:balanced_results_test} presents the new results adapting our loss function.
\begin{table}[h!]
\footnotesize
\begin{center}
\caption{\label{font-table} Comparison between isolated intent (without context) and windows-based approaches using loss values. The highest results are highlighted in red. Baseline results are highlighted in blue.}
\begin{tabular}{ c ||c||c||c||c }
  \hline
    \textbf{Approach}&\textbf{Accuracy}&\textbf{Recall}&\textbf{Precision}&\textbf{F1-Score}\\
  \hline
  \textbf{without context} & \textcolor{blue}{85.19} & \textcolor{blue}{84.62} & \textcolor{blue}{86.17} & \textcolor{blue}{85.37} \\
  \hline
  \textbf{all context} & 70.77 & 66.85 & 79.25 & 72.38 \\
  \hline
  \textbf{user context} & 78.64 & 71.55 & 88.67 & 78.92 \\
  \hline
  \textbf{user-system context} & 86.21 & 83.97 & \textcolor{red}{90.14} & 86.89\\
  \hline
  \textbf{last-user context} &  \textcolor{red}{87.58} & \textcolor{red}{85.96} & 89.49 & \textcolor{red}{87.65} \\
  \hline
  \textbf{last-system context} &87.45 & 85.90 & 89.37 & 87.57 \\
  \hline
\end{tabular}
\label{table:balanced_results_test}
\end{center}
\end{table}

Our current model gets better results using the last utterance from the user, and system \textit{(user-system context, last-user context, and last-system context)}. In the same way, using the \textit{last-user context} to carry out a concatenation with the current utterance contributes to the best context to predict the current user's intent.

\section{CONCLUSIONS AND FUTURE WORKS}
\label{sec:CONCLUSIONS_FUTURE_WORKS}
Intent classification in dialog flow processing can benefit from addressing contextual information. In this work, we proposed windows-based approaches for intent classification tasks. Our window-based strategy made use of distinct types of windows on the contextual information in a conversation between a user and the system. We identified and defined five types of trajectories. Our study experimented with these trajectories. For experimental purposes, we used WvGD dataset, which provided dialogue flows of various sizes between 3 and 15 queries. We conducted experiments to understand the added value of our defined windows-based approaches for intent classification. Our results achieved substantial improvements over the baseline in three approaches using user's utterance and system's response from previous messages as dialogue context. In addition, we modified our loss function to penalize misclassification of the underrepresented classes more than the dominant ones, this enabled us to address the imbalance of our dataset. The final results showed benefits in this decision and increased the accuracy during the classification process. As far as future works are concerned, we are going to investigate the use of our windows-based approach on Dialogue State Tracking task because current works do not standardize the dialogue history, and we are going to propose and experiment with other approaches to intent classification relying on contextual information.

\section{ACKNOWLEDGMENT}
\label{sec:ACKNOWLEDGMENT}

Jeanfranco D. Farfan-Escobedo thanks the financial support of Sinch Latin America.

\bibliographystyle{IEEEtran}
\bibliography{ieee.bib}

\end{document}